# Risk of Bias in Chest Radiography Deep Learning Foundation Models

Ben Glocker* · Charles Jones · Mélanie Roschewitz · Stefan Winzeck

Department of Computing, Imperial College London, London, SW7 2AZ, United Kingdom

*Corresponding author: b.glocker@imperial.ac.uk

## Abstract

### Purpose

To analyze a recently published chest radiography foundation model for the presence of biases that could lead to subgroup performance disparities across biological sex and race.

### Materials and Methods

This retrospective study used 127,118 chest radiographs from 42,884 patients (mean age, 63 ± [SD] 17 years; 23,623 male, 19,261 female) from the CheXpert dataset collected between October 2002 and July 2017. To determine the presence of bias in features generated by a chest radiography foundation model and baseline deep learning model, dimensionality reduction methods together with two-sample Kolmogorov-Smirnov tests were used to detect distribution shifts across sex and race. A comprehensive disease detection performance analysis was then performed to associate any biases in the features to specific disparities in classification performance across patient subgroups.

### Results

Ten out of twelve pairwise comparisons across biological sex and race showed statistically significant differences in the studied foundation model, compared with four significant tests in the baseline model. Significant differences were found between male and female (P < .001) and Asian and Black patients (P < .001) in the feature projections that primarily capture disease. Compared with average model performance across all subgroups, classification performance on the 'no finding' label dropped between 6.8% and 7.8% for female patients, and performance in detecting 'pleural effusion' dropped between 10.7% and 11.6% for Black patients.

### Conclusion

The studied chest radiography foundation model demonstrated racial and sex-related bias leading to disparate performance across patient subgroups and may be unsafe for clinical applications.

# Introduction

Deep learning-based predictive models have found great success in medical imaging applications, such as disease detection in chest radiography (CXR) (1). However, training of these models requires access to large amounts of representative data. Generalization across different clinical sites remains a major challenge for wider clinical adoption (2). Training on limited data makes models susceptible to failure whenever the data characteristics change, often caused by differences in the patient demographics (i.e. population shift) and/or imaging technique (i.e. acquisition shift) (3,4).

Foundation models have emerged as a promising solution to mitigate these issues (5,6). These models are pretrained on large-scale, heterogeneous, and diverse datasets, often by self-supervised or semi-supervised learning strategies that do not require ground truth annotations, with the hope to provide robust 'backbones' for task-agnostic feature extraction. These backbone features then serve as inputs for the subsequent, data-efficient training of task-specific prediction models. The term 'foundation model' is now widely used to describe pretrained, versatile deep learning models that can be adapted to a wide range of downstream prediction tasks (7).

In medical imaging, pretraining is particularly attractive due to the difficulty of collecting large amounts of high-quality training data. Recent work includes self-supervised pretraining on large unlabeled medical imaging datasets, which appears to not only improve performance on data from similar sources (in-distribution) but also across various downstream tasks on new, out-of-distribution data (8). These findings were corroborated by Ghesu et al, (9) who proposed a foundation model trained on over one million diverse medical images. Similarly, Sellergren et al (10) have recently developed a CXR foundation model, demonstrating that it can improve performance in downstream tasks as well as drastically reduce the amount of labeled training data required for task-specific finetuning. The pretrained model yielded an area under the receiver operating characteristic curve (AUC) of 0.95 for detecting tuberculosis when using only 45 chest radiographs for task-specific training, which was noninferior to radiologist performance. Outcome prediction after COVID-19 was found to be better when freezing the backbone versus finetuning the entire model on the complete dataset, when using only 528 chest radiographs for training.

Despite their increasing popularity, little is known about potential biases encoded and reinforced in these foundation models, as well as their effect on embedding biases in downstream models. Previous studies on foundation models in medical imaging largely lack a comprehensive bias analysis. This deserves a closer investigation in light of ethical and regulatory concerns regarding use of foundation models in healthcare applications (11) and, specifically, in radiology (12). Use of foundation models in medical imaging may be of particular concern given the recently demonstrated ability of deep learning models to accurately recognize protected characteristics such as racial identity and other demographic information (13,14).

In this study, we specifically analyze a recently published CXR foundation model proposed in the work by Sellergren et al. (10). We inspect the generated features of this proprietary model for the presence of biases that could potentially lead to disparate performance across patient subgroups (15,16). We conduct a comprehensive subgroup performance analysis when employing the foundation model for the downstream application of disease detection. Our performance analysis associates biases found in feature representation to specific performance disparities in protected subgroups.

## Materials and Methods

This retrospective study is exempt from ethical approval, as the analysis is based on secondary data that is publicly available, and no permission is required to access the data. The study was compliant with the Health Insurance Portability and Accountability Act (HIPAA).

### Study Sample

We used a sample from the publicly available CheXpert dataset (17) composed of data from a total of 42,884 patients with 127,118 chest radiographs divided into three sets for training (76,205 radiographs), validation (12,673 radiographs) and testing (38,240 radiographs) collected between October 2002 and July 2017. The study sample and data splits used in our study are identical to the ones used in the recent study by Gichoya et al. (13). We refer the reader to the study by Gichoya et al, and specifically to their extensive supplementary material (page 2), for an excellent discussion and further information about the definitions of the used racial groupings. The code repository (https://github.com/biomedia-mira/cxr-foundation-bias) released with our study contains detailed information on how to construct the study sample from the original CheXpert dataset.

### Models

The primary model of our investigation is the recently proposed CXR foundation model (10). According to the description, this model was first pretrained on a large corpus of natural images followed by a second pretraining on more than 800,000 chest radiographs from India and the United States. This second pretraining step is based on supervised contrastive learning to specifically train for the classification of images with and without pathology, leveraging disease labels extracted from radiology reports using natural language processing. The foundation model is intended to serve as a robust feature extractor used for subsequent training of downstream, task-specific prediction models. Access to the foundation model is made available through a programming interface which allows only the processing of input images, with the output corresponding to the generated features. The network weights of the CXR foundation model itself, however, are not publicly available, meaning it is not possible to update the parameters of the feature extractor during training of downstream tasks.

In order to compare the CXR foundation model with a 'traditional' approach of model development, we adopted a widely used deep convolutional neural network, DenseNet-121 (18), trained and validated on the CheXpert training and validation sets. This network is pretrained on natural images and then finetuned for the task of disease detection in chest radiography, using the disease annotations available in the CheXpert dataset. We used the identical, publicly available model described in the work by Glocker et al. (19). The already fully trained model was obtained from their code repository, and no further modifications were made. Hereafter, we refer to this baseline model as the CheXpert model. A similar model has been used in other works and is considered state-of-the-art for chest radiography disease detection (16,20).

### Model Inspection

To analyze whether biases may persist in the features generated by the foundation model, we used the CheXpert test set with 38,240 scans. We employed test set resampling to correct for variations across subgroups, such as racial imbalance, differences in age, and varying prevalence of disease. We then used the feature exploration framework proposed by Glocker et al. (19). First, we obtained the corresponding features for the entire test set by passing each scan through the model backbones of the CXR foundation model and the CheXpert model. The high-dimensional feature vectors were then projected down to lower dimensional feature spaces using principal component analysis (PCA) and t-distributed stochastic neighbor embedding (t-SNE). For PCA, these new dimensions (also called modes) capture the direction of the largest variation in the high-dimensional feature space. This means that for a model trained for disease detection, we find the strongest separation of samples with and without the disease in the first few modes of PCA. We applied t-SNE on top of PCA using all modes to retain 99% of the variance, aiming to capture the overall similarity between samples in the original high-dimensional feature space. For the bias analysis, we randomly sampled a set of 3,000 patients (1,000 samples from each racial group) and inspected whether the PCA modes that separate samples by disease may additionally separate non-disease related patient characteristics such as biological sex or racial identity. Similarly, we inspected t-SNE projections to determine whether any groupings or distributional differences appear across patient subgroups. Differences found across subgroups in PCA and/or t-SNE projections may indicate that the underlying features do not only capture variation in disease status but also encode biases with respect to protected patient characteristics.

**Model Performance**

While biases in the features may not necessarily be problematic, it is important to assess if such biases may affect downstream disease detection performance. To this end, we performed a comprehensive subgroup performance analysis comparing different disease detection models built with and without the use of the CXR foundation model. First, we used the CXR foundation model as a feature extractor to build three different disease detection models by training classification submodels with increasing complexity. The classification submodels take the features generated by the CXR foundation model as inputs and produce multi-label, probabilistic outputs for different disease labels. The three submodels correspond to a single, fully-connected classification layer, denoted as CXR-Linear, and two multi-layer perceptrons (MLPs) with three and five hidden layers, denoted as CXR-MLP-3 and CXR-MLP-5, respectively. These disease detection models represent the intended use of the CXR foundation model, acting as a mechanism to facilitate effective transfer learning for task-specific training of prediction models. All three classification models were trained using the CheXpert training set with the corresponding validation set being used for model selection. We then compared the performance of these models to our baseline CheXpert model, a DenseNet-121, trained on the exact same data. All models were then evaluated on the CheXpert test set, using test set resampling to correct for demographic variations across subgroups. Here, we followed the test set resampling strategy for an unbiased estimation of subgroup performance as described by Glocker et al. (19). We used resampling with replacement to construct balanced test sets, correcting for racial imbalance, differences in age, and varying prevalence of disease. In this study, we evaluated and compared disease detection performance on four different labels, 'no finding', 'pleural effusion', 'cardiomegaly', and 'pneumothorax', to provide a variety of results and insights across different target predictions.

**Statistical Analysis**

To determine whether the features generated by a model were biased, we used two-sample Kolmogorov-Smirnov tests to determine p-values for the null hypothesis that the marginal distributions for a given pair of subgroups are identical in each of the first four modes of PCA. These statistical tests were performed for all relevant pairwise comparisons regarding the presence of disease, biological sex, and race. The p-values were adjusted for multiple testing using the Benjamini-Yekutieli procedure, and significance was determined at a 95% confidence level ($P < .05$).

To evaluate and compare the disease detection performance of different models, we computed the AUC, true positive rate (TPR), and false positive rate (FPR). TPR and FPR in subgroups were determined at a fixed decision threshold, which was optimized for each model to yield an FPR of 0.20 on the whole patient sample. The fixed target FPR allows for immediate identification of performance deviations across subgroups. To provide a single measure of classification performance, we report the Youden's J statistic (measured at the target FPR), which is defined as J = TPR - FPR. We used bootstrapping with 2,000 samples to calculate 95% confidence intervals.

All information to recreate the exact study sample used in this paper, including splits of training, validation, and test sets, and all code that is required for reproducing the results are available under an open-source Apache 2.0 license in a dedicated GitHub repository (https://github.com/biomedia-mira/cxr-foundation-bias). All deep learning models were implemented in PyTorch. The model inspection via PCA and t-SNE was done using scitkit-learn. All statistical tests were performed using SciPy version 1.10.0.

## Results

### Patient Characteristics

The study included 127,118 chest radiographs from 42,884 Asian, Black, and White patients (mean age, 63 ± [SD] 17 years; 23,623 male, 19,261 female). Table 1 provides a full breakdown of the study sample characteristics.

### Model Inspection

Figure 1 presents the PCA-based feature space analysis of the two inspected backbone models, showing marginal distributions for different subgroups across the first four PCA modes. The corresponding scatter plots are given in the Supplementary Material in Figure S1. Visually, we observed more and larger differences in the marginal distributions for the CXR foundation model across the protected characteristics of biological sex and race. This is particularly visible in the subgroup distributions for biological sex (second column in Fig. 1) where clear shifts between males and females were observed in all four PCA modes, while no obvious separation is visible for the CheXpert model. Similarly, we observed larger differences in the distributions of racial groups in the CXR foundation model compared with the model trained on CheXpert (third column in Fig. 1). Figure 2 presents the marginal distributions for the t-SNE projections, with the corresponding scatter plots shown in Figure S2. As the orientation of t-SNE dimensions is somewhat arbitrary, it was generally more difficult to visually detect any potential relationship between disease information and protected characteristics. However, we still observed larger differences between subgroup distributions for both biological sex and race for the CXR foundation model compared with the CheXpert model, which was visible when focusing on the marginal distributions in the second and third column of Figure 2.

The statistical analysis confirmed these qualitative observations (Table 2). For biological sex, we found significant differences between the marginal distributions for male and female patients in all four PCA modes ($P < .001$, $P = .0013$, $P < .001$, $P < .001$), compared with no evidence of differences found in the CheXpert model ($P > .99$, $P = .26$, $P > .99$, $P = .15$). Significant differences are also found between the groups of Asian and Black patients in all four PCA modes in the CXR foundation model (all $P < .001$) versus two significant differences in the first and second mode of PCA for the CheXpert model ($P = .021$, $P < .001$, $P = .29$, $P = .40$). More differences were also observed between White and Asian and White and Black patients in the CXR foundation model compared to the CheXpert model. Focusing on the first three PCA modes, which primarily capture differences in the features related to presence of disease (indicated by the significant differences between 'no finding' and 'pleural effusion'), we found that ten out of twelve pairwise comparisons on protected characteristics of biological sex and race showed significant differences in the CXR foundation model, compared with four out of twelve significant tests in the CheXpert model. Considering the explained variance for each PCA mode (see Table 2), we found that the first three PCA modes combined explained more than 53% of the variance in the CheXpert model compared with 37% in the CXR foundation model, indicating that the latter captures substantially more information in its feature representation that may be unrelated to disease prediction. To rule out within patient cluster effects due to the presence of multiple scans per patient within the test set, we redid the analysis for a subsampled test set with only one scan per patient. The overall findings and conclusions remained unchanged, confirming the larger disparities for the CXR foundation model.

**Model Performance**

The differences in performance in terms of Youden's J statistic across models and patient subgroups are summarized in the graphs presented in Figure 3. We found that the models built on top of the CXR foundation model, CXR-Linear, CXR-MLP-3, and CXR-MLP-5, consistently underperformed compared with the CheXpert model. Compared with average model performance across all subgroups, performance in detecting 'no finding' dropped between 6.8% and 7.8% for female patients, and performance in detecting 'pleural effusion' dropped between 10.7% and 11.6% for Black patients. We also observed a drastic decrease in overall performance in classifying 'cardiomegaly' across all patient groups. Additionally, we observed a clear difference in relative performance, leading to concerning subgroup disparities. Figure 4 presents the relative change in performance in terms of Youden's J statistic for each subgroup when compared with each model's average performance over all subgroups. We observed substantially larger disparities in relative performance across biological sex and race for the CXR foundation models compared with the CheXpert model. The absolute and relative performances in terms of AUC are summarized in the graphs presented in Figures S3 and S4, with similar findings of larger performance disparities across subgroups for the CXR foundation model. Detailed results of the subgroup performance analysis with various performance metrics are given in Tables S1-S4 in the Supplementary Material.

## Discussion

This investigation aims to highlight the potential risks for using foundation models in the development of medical imaging artificial intelligence. The fact that the investigated CXR foundation model encodes protected characteristics more strongly than a task-specific backbone raises concerns, as these biases could amplify already existing health disparities (21–24). Our bias analysis showed significant differences between features related to disease detection across biological sex ($P < .001$) and race ($P < .001$). When using the foundation model in downstream disease detection, our subgroup performance analysis revealed a substantial degradation in classification performance, with specific disparities in protected subgroups. Classification performance on the 'no finding' label dropped between 6.8% and 7.8% for female patients, and performance in detecting 'pleural effusion' dropped between 10.7% and 11.6% for Black patients.

These findings are in line with the study by Seyyed-Kalantari et al who found performance disparities in chest radiograph disease detection across underrepresented subgroups (16). Our results indicate that there is a risk of bias for classification models built on top of features extracted with the CXR foundation model. Identifying these issues for the CheXpert dataset is noteworthy, as this dataset was specifically used in the original study to evaluate the generalization ability of the foundation model (10). This highlights that even for artificial intelligence developers, it remains difficult to assess whether their models may be suitable for a specific target dataset. End-users of third-party foundation models, who may have less knowledge and insights about model pretraining, may find it even more difficult to assess risk of bias for their specific application and data. This is particularly concerning in the light of recent studies demonstrating that medical images encode protected characteristics that can be recognized by deep learning models (13,14). The availability of diverse and representative datasets with detailed demographic information will be key for algorithmic auditing and comprehensive assessment of algorithmic bias (25–27).

We believe that our findings have implications beyond the studied model, as the difficulty to scrutinize foundation models applies in general, as pointed out in the work by Bommasani et al. (5). This difficulty stems from the fact that detailed information about the data-generating processes and the exact training strategies is often missing. If biases remain undetected, they can cause serious harm, such as underdiagnosis of underserved populations (16). To mitigate these risks, it will be important to better understand how biases are encoded and how we may prevent the use of undesired information in prediction tasks (19,28). Of note, when a prediction model is trained via finetuning of a pretrained foundation model, we typically have two options: (a) we can 'unfreeze' the backbone model, allowing the finetuning process to modify the mapping from input images to features, potentially overriding biases in the backbone; or (b) we can 'freeze' the backbone model and learn only the parameters specific to the task prediction model, which usually requires substantially fewer training data, and is therefore more appealing in practice. Arguably, the latter is more likely to carry forward any biases from the backbone, as the method that features are generated remains unchanged. If the backbone features separate patients based on a protected characteristic, it is likely that the task-specific prediction model will learn separate mechanisms for the different subgroups. Finetuning is unlikely to be able to 'unlearn' biases and may potentially exploit shortcuts for making predictions due to the presence of undesirable correlations in the training data (29,30). In this context, the provided access to the CXR foundation model by Sellergren et al. (10) is problematic, as the developers only offer option (b). The original backbone model is not publicly shared; hence, one cannot update the mechanism for feature extraction when performing the task-specific finetuning, which limits the use of debiasing techniques (28,31). The observed differences in absolute and relative subgroup performance may be partly explained by the fact that the CXR foundation model was frozen during training of the classification submodels.

Our study has important limitations. We analyzed only one foundation model that was trained in a specific way, using supervised contrastive learning. Future work should explore whether biases also manifest in other chest radiography foundation models that are trained differently, such as via self-supervision without any annotations (8,9). Such models, however, are currently not publicly available. We may expect to find similar biases in models with fully self-supervised training, as such training strategies encourage grouping of individuals in feature spaces that are visually similar. Therefore, we may expect to see clusters for biological sex and potentially race, as these characteristics are known to be separable with high predictive accuracy (13,14,32). We believe that our work may provide a methodological basis for future bias analyses. Another limitation is that we could not shed light on the exact origin of the biases in the studied foundation model due to insufficient insights into the exact training data characteristics. While the amount of training data for the foundation model was considerably large, with more than 800,000 chest radiographs, it was limited to data from two countries, India and the United States. Most images, over 700,000, were reported to come from India, which may contribute to the observed bias across racial subgroups. It has been argued previously that mixing effects from dataset bias are notoriously difficult to analyze (33). Thus, a more systematic approach with controlled, simulated environments to specifically inject different types of bias may be required to isolate the effect of each bias on classification performance.

In conclusion, our study demonstrates that biases in the CXR foundation model related to race and biologic sex led to substantial performance disparities across protected subgroups. To minimize the risk of bias associated with use of foundation models in critical applications such as clinical decision making, we argue that these models need to be fully accessible and transparent. This is important for allowing a more detailed analysis of potential biases and scrutiny of the resulting task-specific prediction models. Here, we advocate for comprehensive bias analysis and subgroup performance analysis to become integral parts in the development and auditing of future foundation models, which is essential for their safe and ethical use in healthcare applications.

## Data and code availability

All data and code used in this work is publicly available. The CheXpert imaging dataset can be obtained from https://stanfordmlgroup.github.io/competitions/chexpert/. The demographic information is available on https://stanfordaimi.azurewebsites.net/datasets/192ada7c-4d43-466e-b8bb-b81992bb80cf.

All information and code to reproduce our results and recreate the exact study sample used in this paper including splits of training, validation, and test sets is available under an open source Apache 2.0 license in our dedicated GitHub repository https://github.com/biomedia-mira/cxr-foundation-bias/.

## Acknowledgments

B.G. received funding from the European Research Council (ERC) under the European Union's Horizon 2020 research and innovation programme (Grant Agreement No. 757173, Project MIRA). S.W. is supported by the UKRI London Medical Imaging & Artificial Intelligence Centre for Value Based Healthcare. C.J. is supported by Microsoft Research and EPSRC. M.R. is funded through an Imperial College London President's PhD Scholarship.

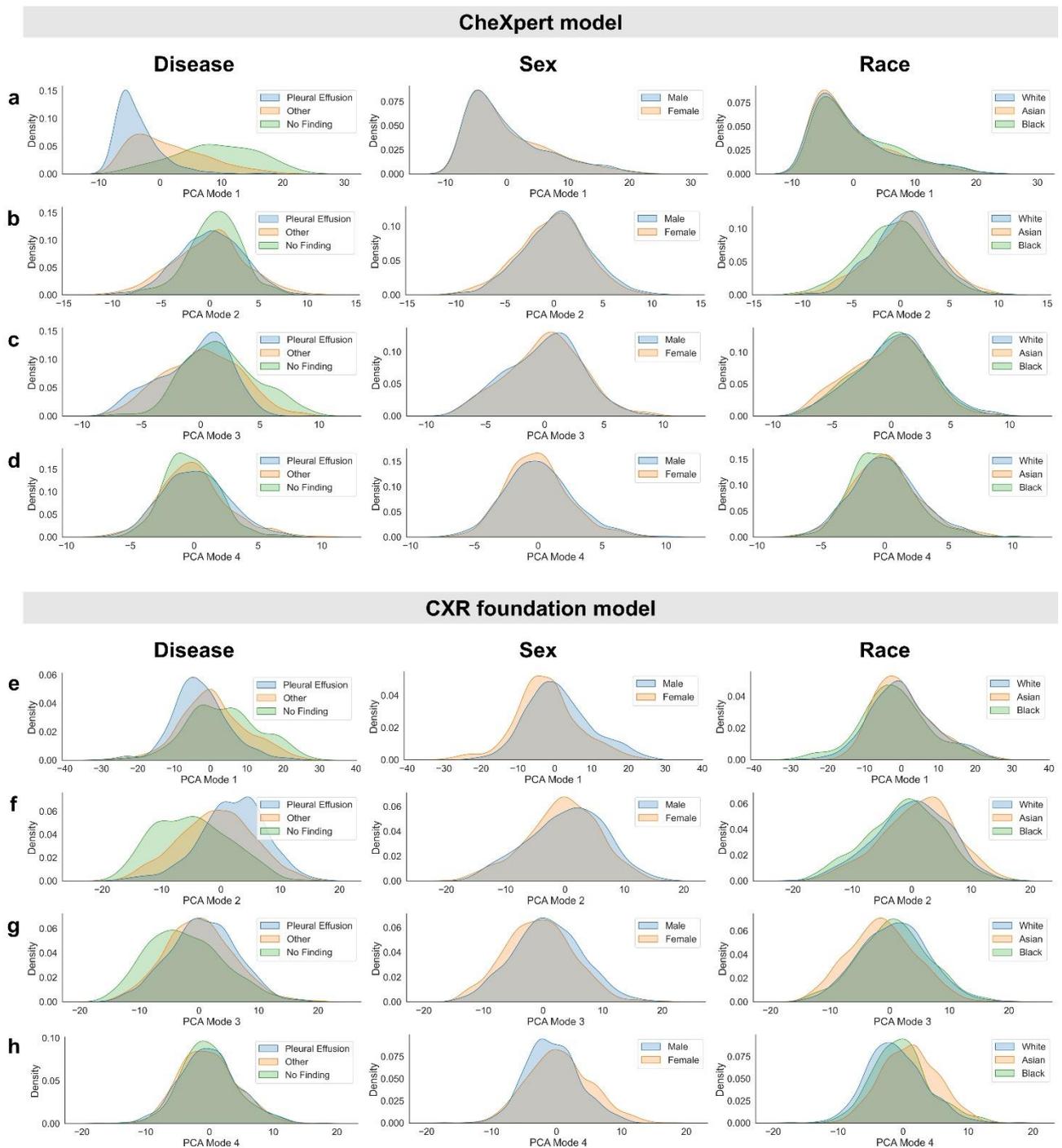

Figure 1. **Inspection of subgroup distribution shifts in the PCA feature space projections.** Marginal distributions are plotted across subgroups for the first four modes of PCA applied to the extracted feature vectors of the CheXpert test data for (a-d) the CheXpert model and (e-h) the CXR foundation model. The plots are generated using a random set of 3,000 patients (1,000 samples from each racial group). Marginal distributions are normalized independently to remove differences in subgroup base rates and shown for different characteristics (from left to right): presence of disease, biological sex, and racial identity. Larger distribution shifts across sex and race are observed for the CXR foundation model. CXR = chest radiography, PCA = principal component analysis

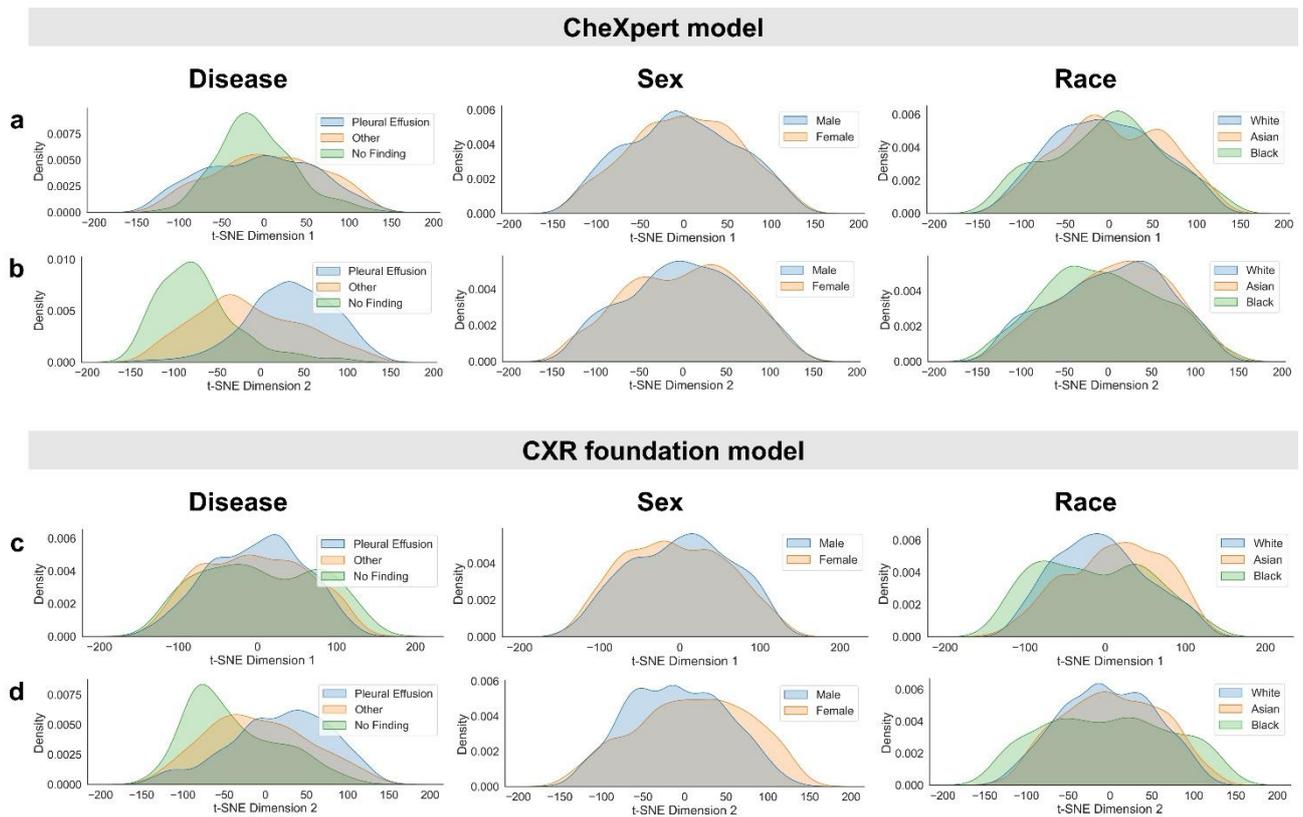

Figure 2. **Inspection of subgroup distribution shifts in the t-SNE feature space projections.** Marginal distributions are plotted across subgroups for the two dimensions of t-SNE applied to the extracted feature vectors of the CheXpert test data for (a,b) the CheXpert model and (c,d) the CXR foundation model. The plots are generated using a random set of 3,000 patients (1,000 samples from each racial group). Marginal distributions are normalized independently to remove differences in subgroup base rates and shown for different characteristics (from left to right): presence of disease, biological sex, and racial identity. Larger distribution shifts across sex and race are observed for the CXR foundation model. CXR = chest radiography, t-SNE = t-distributed stochastic neighbor embedding

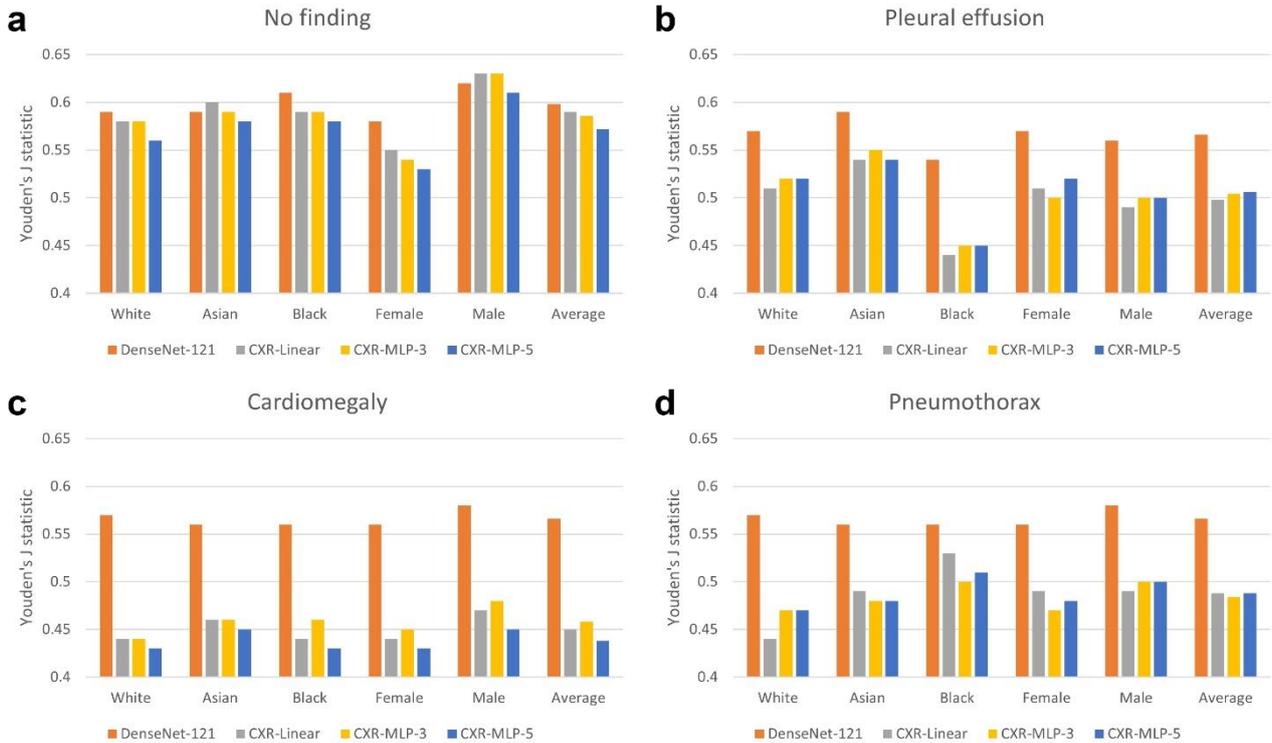

Figure 3. **Comparison of disease detection performance across patient subgroups.** Average classification performance across patient subgroups is shown in terms of Youden's J statistic for the DenseNet-121 CheXpert model and three variants of the CXR foundation model. Classification performance is shown on four different labels of (a) 'no finding', (b) 'pleural effusion', (c) 'cardiomegaly', and (d) 'pneumothorax'. The CXR foundation models consistently underperformed compared with the CheXpert model, with specific underperformance on the subgroup of female patients for 'no finding' and the subgroup of Black patients on 'pleural effusion'. There was also a drastic decrease in overall performance across all subgroups for the CXR foundation models for 'cardiomegaly'. CXR = chest radiography, MLP = multi-layer perceptrons

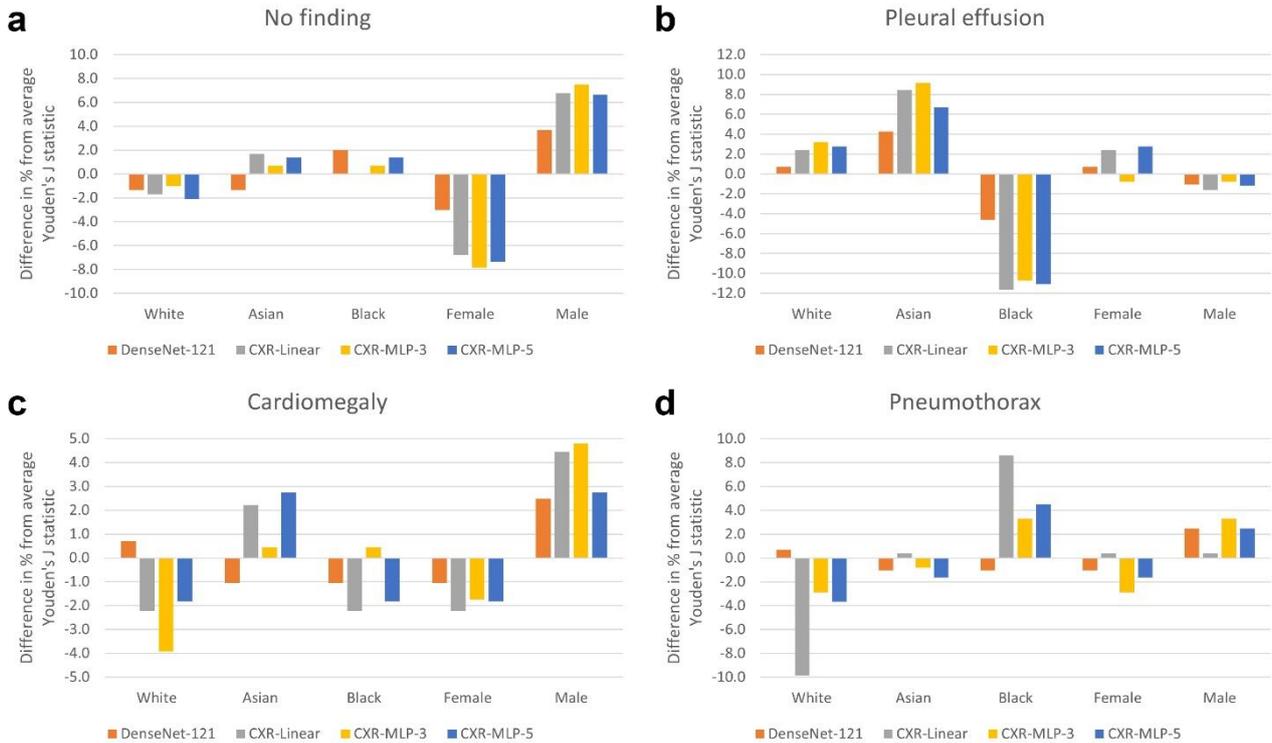

Figure 4. **Relative change in disease detection performance across patient subgroups.** The relative change in performance for each subgroup was measured by comparing the subgroup performance with each model's average performance over all subgroups. Performance is measured in terms of Youden's J statistic on the labels of (a) 'no finding', (b) 'pleural effusion', (c) 'cardiomegaly', and (d) 'pneumothorax'. There were substantially larger disparities in relative performance across biological sex and race for the three CXR foundation models, CXR-Linear, CXR-MLP-3, and CXR-MLP-5 when compared with the DenseNet-121 CheXpert model. CXR = chest radiography, MLP = multi-layer perceptrons

Table 1. Characteristics of the Study Sample

| | CheXpert Dataset | | | | | |
|---|---|---|---|---|---|---|
| | All | White | Asian | Black | Male | Female |
| Attribute | All data | | | | | |
| Patients | 42,884 | 33,338 | 6,642 | 2,904 | 23,623 | 19,261 |
| Scans | 127,118 | 99,027 (78) | 18,830 (15) | 9,261 (7) | 74,682 (59) | 52,436 (41) |
| Age (years) | 63 ± 17 | 64 ± 17 | 61 ± 17 | 56 ± 17 | 62 ± 17 | 64 ± 18 |
| Female | 52,436 (41) | 39,735 (40) | 8,132 (43) | 4,569 (49) | - | - |
| No finding | 10,916 (9) | 8,236 (8) | 1,716 (9) | 964 (10) | 6,280 (8) | 4,636 (9) |
| Pleural effusion | 51,574 (41) | 40,545 (41) | 7,953 (42) | 3,076 (33) | 30,171 (40) | 21,403 (41) |
| Cardiomegaly | 15,790 (12) | 11,561 (12) | 2,375 (13) | 1,854 (20) | 9,793 (13) | 5,997 (11) |
| Pneumothorax | 12,237 (10) | 9,653 (10) | 2,013 (11) | 571 (6) | 7,280 (10) | 4,957 (9) |
| | Training data | | | | | |
| Patients | 25,730 | 20,034 | 3,945 | 1,751 | 14,164 | 11,566 |
| Scans | 76,205 | 59,238 (78) | 11,371 (15) | 5,596 (7) | 44,773 (59) | 31,432 (41) |
| Age (years) | 63 ± 17 | 64 ± 17 | 62 ± 17 | 56 ± 17 | 62 ± 17 | 64 ± 18 |
| Female | 31,432 (41) | 23,715 (40) | 4,976 (44) | 2,741 (49) | - | - |
| No finding | 6,514 (9) | 4,910 (8) | 1,046 (9) | 558 (10) | 3,756 (8) | 2,758 (9) |
| Pleural effusion | 31,015 (41) | 24,405 (41) | 4,754 (42) | 1,856 (33) | 18,174 (41) | 12,841 (41) |
| Cardiomegaly | 9,353 (12) | 6,835 (12) | 1,397 (12) | 1,121 (20) | 5,864 (13) | 3,489 (11) |
| Pneumothorax | 7,435 (10) | 5,871 (10) | 1,235 (11) | 329 (6) | 4,461 (10) | 2,974 (9) |
| | Validation data | | | | | |
| Patients | 4,288 | 3,348 | 666 | 274 | 2,367 | 1,921 |
| Scans | 12,673 | 9,945 (79) | 1,809 (14) | 919 (7) | 7,643 (60) | 5,030 (40) |
| Age (years) | 62 ±17 | 63 ± 17 | 62 ± 17 | 55 ± 16 | 62 ± 17 | 63 ± 18 |
| Female | 5,030 (40) | 3,933 (40) | 667 (37) | 430 (47) | - | - |
| No finding | 1,086 (9) | 817 (8) | 175 (10) | 94 (10) | 602 (8) | 484 (10) |
| Pleural effusion | 5,049 (40) | 3,988 (40) | 738 (41) | 323 (35) | 3,086 (40) | 1,963 (39) |
| Cardiomegaly | 1,548 (12) | 1,156 (12) | 222 (12) | 170 (18) | 987 (13) | 561 (11) |
| Pneumothorax | 1,220 (10) | 956 (10) | 168 (9) | 96 (10) | 744 10) | 476 (9) |
| | Test data | | | | | |
| Patients | 12,866 | 9,956 | 2,031 | 879 | 7,092 | 5,774 |
| Scans | 38,240 | 29,844 (78) | 5,650 (15) | 2,746 (7) | 22,266 (58) | 15,974 (42) |
| Age (years) | 63 ± 17 | 64 ± 17 | 61 ± 17 | 57 ± 16 | 63 ± 16 | 64 ± 18 |
| Female | 15,974 (42) | 12,087 (41) | 2,489 (44) | 1,348 (49) | - | - |
| No finding | 3,316 (9) | 2,509 (8) | 495 (9) | 312 (11) | 1,922 (9) | 1,394 (9) |
| Pleural effusion | 15,510 (41) | 12,152 (41) | 2,461 (44) | 897 (33) | 8,911 (40) | 6,599 (41) |
| Cardiomegaly | 4,889 (13) | 3,570 (12) | 756 (13) | 563 (21) | 2,942 (13) | 1,947 (12) |
| Pneumothorax | 3,582 (9) | 2,826 (9) | 610 (11) | 146 (5) | 2,075 (9) | 1,507 (9) |

Note.—Breakdown of demographics over the set of patient scans by racial groups and training, validation and test splits. Percentages in brackets are with respect to the number of scans. Age is reported as mean ± SD. We also report the number of unique patients.

Table 2. Kolmogorov-Smirnov Tests for Comparing Marginal Distributions Across PCA Modes

| | | **CheXpert model** | | | | |
|---|---|---|---|---|---|---|
| | | No finding / Pleural effusion | White / Asian | Asian / Black | Black / White | Male / Female |
| Mode | Exp. Var. | | | p-values | | |
| PCA mode 1 | 31.2% | <0.0001** | 1.00 | 0.021* | 0.40 | 1.00 |
| PCA mode 2 | 9.0% | 0.012* | 1.00 | <0.0001** | <0.0001** | 0.26 |
| PCA mode 3 | 7.9% | <0.0001** | 0.025* | 0.29 | 1.00 | 1.00 |
| PCA mode 4 | 5.1% | 0.12 | 1.00 | 0.40 | 0.16 | 0.15 |
| | | **CXR foundation model** | | | | |
| | | No finding / Pleural effusion | White / Asian | Asian / Black | Black / White | Male / Female |
| Mode | Exp. Var. | | | p-values | | |
| PCA mode 1 | 16.8% | <0.0001** | 0.96 | <0.0001** | 0.00031** | <0.0001** |
| PCA mode 2 | 8.6% | <0.0001** | 0.0013* | <0.0001** | 0.0044* | 0.0013* |
| PCA mode 3 | 7.3% | <0.0001** | <0.0001** | <0.0001** | 1.00 | <0.0001** |
| PCA mode 4 | 4.2% | 1.00 | <0.0001** | <0.0001** | 0.00018** | <0.0001** |

Note.—Two-sample Kolmogorov-Smirnov tests were performed between the pairs of subgroups indicated in each column. The p-values are adjusted for multiple testing using the Benjamini-Yekutieli procedure, and significance is determined at a 95% confidence level. Statistically significant results are marked with * P < .05 and ** P < .001. CXR = chest radiography, PCA = principal component analysis

# Supplementary Material

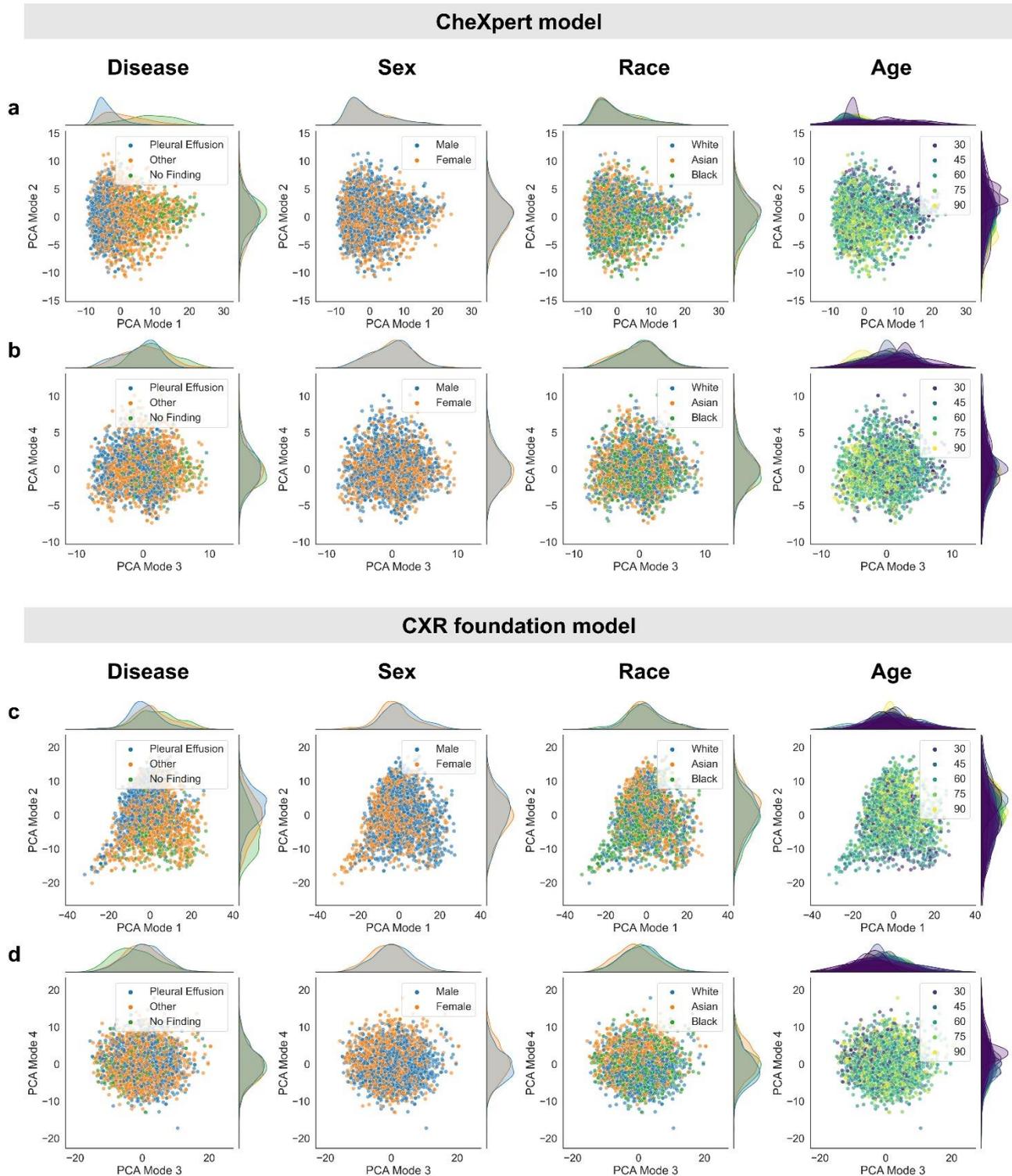

Figure S1. **Inspection of the PCA feature space projections of two different models.** Scatter plots with marginal distributions displayed at the axes are shown for the first four modes of PCA applied to the extracted feature vectors of the CheXpert test data in for (a,b) the CheXpert model, and in for (c,d) the CXR foundation model. The plots are generated using a random set of 3,000 patients (1,000 samples from each racial group). Marginal distributions are normalized independently to remove differences in subgroup base rates. Different characteristics are overlaid in color (from left to right): presence of disease, biological sex, racial identity, and age. Larger distribution shifts across sex and race are observed for the CXR foundation model. CXR = chest radiography, PCA = principal component analysis

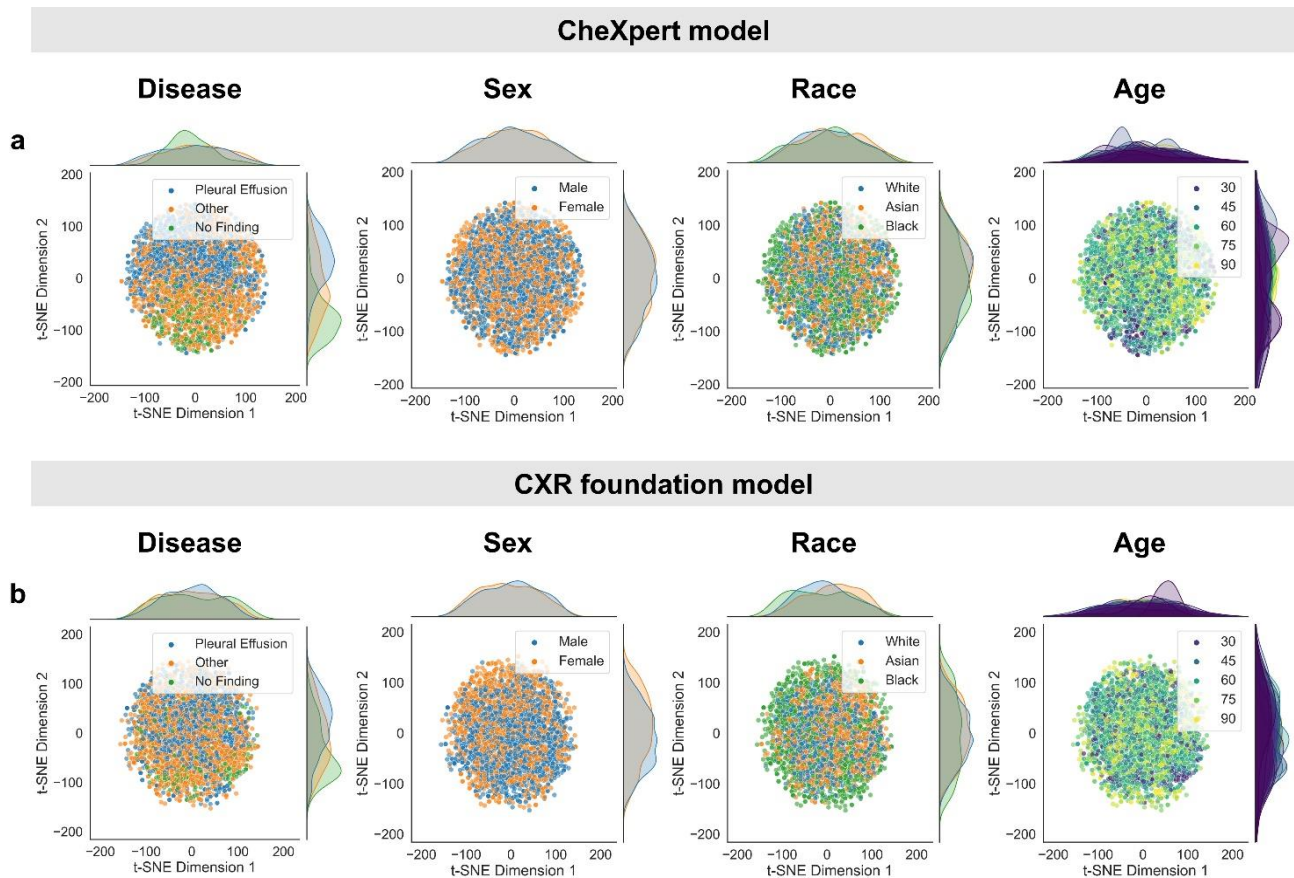

Figure S2. **Inspection of the t-SNE feature space projections of two different models.** Scatter plots with marginal distributions displayed at the axes are shown for the two dimensions of t-SNE applied to the extracted feature vectors of the CheXpert test data in (a) for the CheXpert model, and in (b) for the CXR foundation model. The plots are generated using a random set of 3,000 patients (1,000 samples from each racial group). Marginal distributions are normalized independently to remove differences in subgroup base rates. Different characteristics are overlaid in color (from left to right): presence of disease, biological sex, racial identity, and age. Larger distribution shifts across sex and race are observed for the CXR foundation model. CXR = chest radiography, t-SNE = t-distributed stochastic neighbor embedding

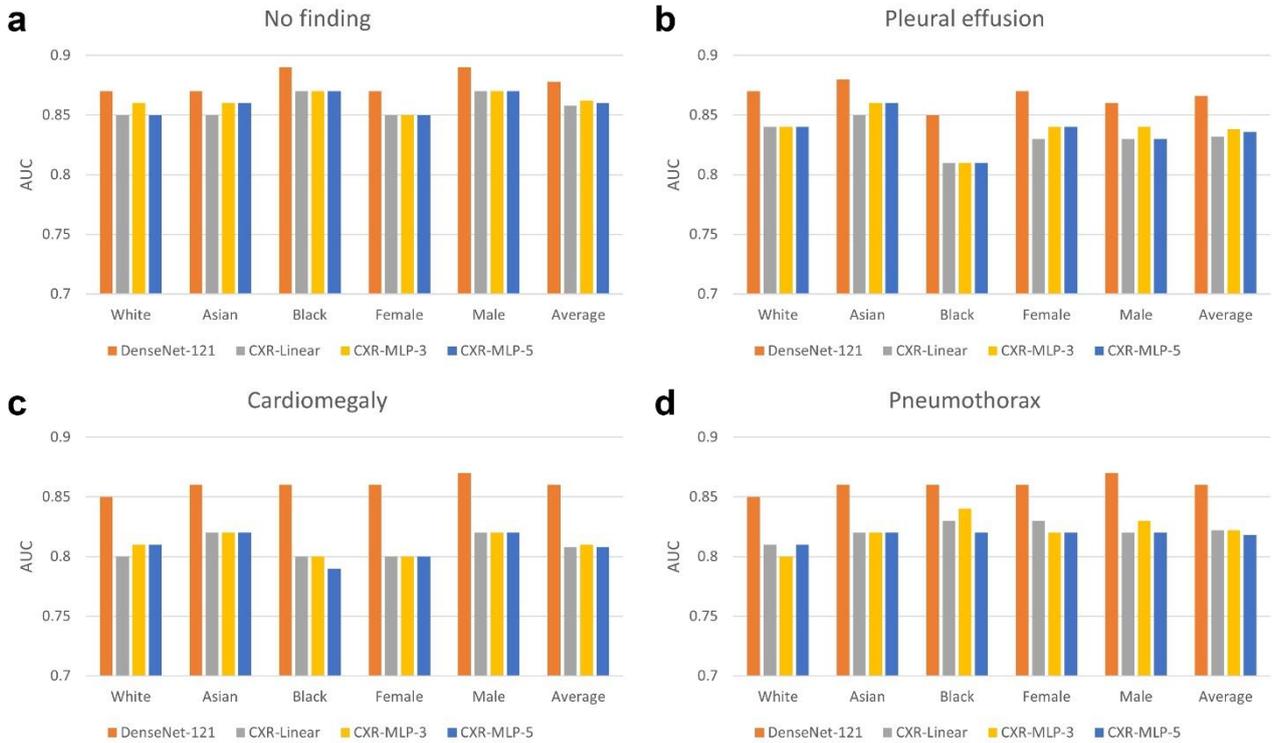

Figure S3. **Comparison of disease detection performance across patient subgroups.** Average classification performance across patient subgroups is shown in terms of AUC for the DenseNet-121 CheXpert model, and three variants of the CXR foundation model. Classification performance is shown on four different labels of (a) 'no finding', (b) 'pleural effusion', (c) 'cardiomegaly', and (d) 'pneumothorax'. The CXR foundation models consistently underperform compared to the CheXpert model. CXR = chest radiography, MLP = multi-layer perceptrons, AUC = area under the receiver operating characteristic curve

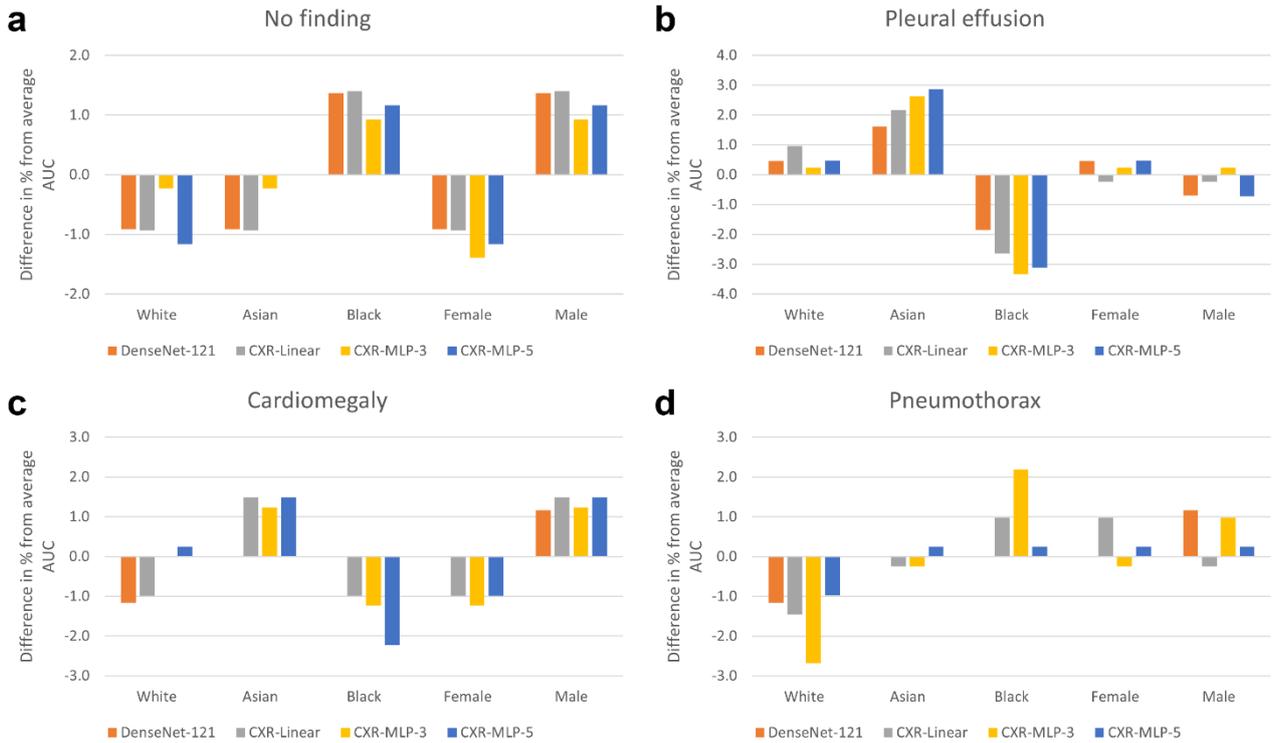

Figure S4. **Relative change in disease detection performance across patient subgroups.** The relative change in performance for each subgroup is measured by comparing the subgroup performance to each model's average performance over all subgroups. Performance is measured in terms of AUC on the labels of (a) 'no finding', (b) 'pleural effusion', (c) 'cardiomegaly', and (d) 'pneumothorax'. We observe substantially larger disparities in relative performance across biological sex and race for the three CXR foundation models, CXR-Linear, CXR-MLP-3, and CXR-MLP-5 when compared to the DenseNet-121 CheXpert model. CXR = chest radiography, MLP = multi-layer perceptrons, AUC = area under the receiver operating characteristic curve

Table S1. Comparison of disease detection performance for the label 'no finding'

| | No finding | | | | |
|---|---|---|---|---|---|
| | White | Asian | Black | Female | Male |
| Model | AUC (95% CI) | | | | |
| DenseNet-121 | 0.87 (0.86-0.88) | 0.87 (0.87-0.88) | 0.89 (0.88-0.89) | 0.87 (0.86-0.87) | 0.89 (0.88-0.89) |
| CXR-Linear | 0.85 (0.85-0.86) | 0.85 (0.85-0.86) | 0.87 (0.87-0.88) | 0.85 (0.84-0.85) | 0.87 (0.86-0.87) |
| CXR-MLP-3 | 0.86 (0.85-0.87) | 0.86 (0.85-0.86) | 0.87 (0.87-0.88) | 0.85 (0.85-0.86) | 0.87 (0.87-0.88) |
| CXR-MLP-5 | 0.85 (0.85-0.86) | 0.86 (0.85-0.86) | 0.87 (0.86-0.88) | 0.85 (0.84-0.85) | 0.87 (0.87-0.88) |
| | TPR (95% CI) | | | | |
| DenseNet-121 | 0.80 (0.78-0.81) | 0.79 (0.78-0.81) | 0.81 (0.80-0.82) | 0.78 (0.76-0.79) | 0.82 (0.81-0.83) |
| CXR-Linear | 0.78 (0.77-0.80) | 0.79 (0.78-0.81) | 0.80 (0.78-0.81) | 0.76 (0.75-0.77) | 0.82 (0.81-0.83) |
| CXR-MLP-3 | 0.78 (0.77-0.79) | 0.79 (0.77-0.80) | 0.80 (0.78-0.81) | 0.74 (0.73-0.75) | 0.83 (0.82-0.84) |
| CXR-MLP-5 | 0.77 (0.75-0.78) | 0.78 (0.77-0.79) | 0.77 (0.76-0.79) | 0.73 (0.71-0.74) | 0.81 (0.80-0.82) |
| | FPR (95% CI) | | | | |
| DenseNet-121 | 0.20 (0.20-0.21) | 0.20 (0.20-0.20) | 0.20 (0.19-0.20) | 0.20 (0.20-0.20) | 0.20 (0.20-0.20) |
| CXR-Linear | 0.20 (0.20-0.20) | 0.19 (0.19-0.20) | 0.21 (0.20-0.21) | 0.21 (0.21-0.21) | 0.19 (0.19-0.20) |
| CXR-MLP-3 | 0.20 (0.19-0.20) | 0.20 (0.20-0.20) | 0.21 (0.20-0.21) | 0.20 (0.20-0.20) | 0.20 (0.20-0.20) |
| CXR-MLP-5 | 0.21 (0.20-0.21) | 0.20 (0.19-0.20) | 0.20 (0.19-0.20) | 0.20 (0.19-0.20) | 0.20 (0.20-0.21) |
| | Youden's J statistic (95% CI) | | | | |
| DenseNet-121 | 0.59 (0.58-0.61) | 0.59 (0.58-0.61) | 0.61 (0.60-0.63) | 0.58 (0.56-0.59) | 0.62 (0.61-0.63) |
| CXR-Linear | 0.58 (0.57-0.60) | 0.60 (0.59-0.62) | 0.59 (0.58-0.61) | 0.55 (0.54-0.56) | 0.63 (0.61-0.64) |
| CXR-MLP-3 | 0.58 (0.57-0.60) | 0.59 (0.57-0.60) | 0.59 (0.58-0.61) | 0.54 (0.53-0.55) | 0.63 (0.62-0.64) |
| CXR-MLP-5 | 0.56 (0.55-0.58) | 0.58 (0.57-0.60) | 0.58 (0.56-0.59) | 0.53 (0.52-0.54) | 0.61 (0.60-0.62) |

Note.—Disease detection results reported separately for each race group and biological sex. TPR and FPR in subgroups are determined using a fixed decision threshold optimized over the whole patient population for a target FPR of 0.20. Performance is reported as mean and 95% confidence interval (CI). AUC = area under the receiver operating characteristic curve, TPR = true positive rate, FPR = false positive rate, CXR = chest radiography, MLP = multi-layer perceptrons

Table S2. Comparison of disease detection performance for the label 'pleural effusion'

| | Pleural effusion | | | | |
|---|---|---|---|---|---|
| | White | Asian | Black | Female | Male |
| Model | AUC (95% CI) | | | | |
| DenseNet-121 | 0.87 (0.86-0.87) | 0.88 (0.88-0.89) | 0.85 (0.84-0.85) | 0.87 (0.87-0.87) | 0.86 (0.86-0.86) |
| CXR-Linear | 0.84 (0.84-0.84) | 0.85 (0.85-0.86) | 0.81 (0.80-0.81) | 0.83 (0.83-0.84) | 0.83 (0.83-0.84) |
| CXR-MLP-3 | 0.84 (0.84-0.85) | 0.86 (0.85-0.86) | 0.81 (0.81-0.81) | 0.84 (0.83-0.84) | 0.84 (0.83-0.84) |
| CXR-MLP-5 | 0.84 (0.84-0.85) | 0.86 (0.85-0.86) | 0.81 (0.80-0.81) | 0.84 (0.83-0.84) | 0.83 (0.83-0.84) |
| | TPR (95% CI) | | | | |
| DenseNet-121 | 0.78 (0.78-0.79) | 0.80 (0.80-0.81) | 0.72 (0.71-0.73) | 0.78 (0.77-0.79) | 0.76 (0.75-0.76) |
| CXR-Linear | 0.73 (0.72-0.74) | 0.76 (0.75-0.77) | 0.62 (0.61-0.63) | 0.73 (0.72-0.74) | 0.68 (0.67-0.69) |
| CXR-MLP-3 | 0.73 (0.72-0.74) | 0.76 (0.75-0.77) | 0.62 (0.61-0.63) | 0.71 (0.70-0.72) | 0.70 (0.69-0.70) |
| CXR-MLP-5 | 0.74 (0.73-0.75) | 0.75 (0.74-0.76) | 0.63 (0.62-0.64) | 0.72 (0.72-0.73) | 0.69 (0.68-0.70) |
| | FPR (95% CI) | | | | |
| DenseNet-121 | 0.21 (0.21-0.21) | 0.21 (0.20-0.21) | 0.18 (0.18-0.19) | 0.20 (0.20-0.21) | 0.20 (0.19-0.20) |
| CXR-Linear | 0.21 (0.21-0.22) | 0.21 (0.21-0.22) | 0.17 (0.17-0.18) | 0.22 (0.21-0.22) | 0.19 (0.18-0.19) |
| CXR-MLP-3 | 0.22 (0.21-0.22) | 0.21 (0.21-0.22) | 0.17 (0.17-0.17) | 0.21 (0.20-0.21) | 0.19 (0.19-0.20) |
| CXR-MLP-5 | 0.21 (0.21-0.22) | 0.21 (0.20-0.21) | 0.18 (0.17-0.18) | 0.21 (0.20-0.21) | 0.19 (0.19-0.20) |
| | Youden's J statistic (95% CI) | | | | |
| DenseNet-121 | 0.57 (0.56-0.58) | 0.59 (0.59-0.60) | 0.54 (0.52-0.55) | 0.57 (0.57-0.58) | 0.56 (0.55-0.57) |
| CXR-Linear | 0.51 (0.51-0.52) | 0.54 (0.54-0.55) | 0.44 (0.44-0.46) | 0.51 (0.50-0.52) | 0.49 (0.48-0.50) |
| CXR-MLP-3 | 0.52 (0.51-0.53) | 0.55 (0.54-0.56) | 0.45 (0.44-0.46) | 0.50 (0.49-0.51) | 0.50 (0.50-0.51) |
| CXR-MLP-5 | 0.52 (0.52-0.53) | 0.54 (0.53-0.55) | 0.45 (0.44-0.46) | 0.52 (0.51-0.52) | 0.50 (0.49-0.50) |

Note.—Disease detection results reported separately for each race group and biological sex. TPR and FPR in subgroups are determined using a fixed decision threshold optimized over the whole patient population for a target FPR of 0.20. Performance is reported as mean and 95% confidence interval (CI). AUC = area under the receiver operating characteristic curve, TPR = true positive rate, FPR = false positive rate, CXR = chest radiography, MLP = multi-layer perceptrons

Table S3. Comparison of disease detection performance for the label 'cardiomegaly'

| | Cardiomegaly | | | | |
|---|---|---|---|---|---|
| | White | Asian | Black | Female | Male |
| Model | AUC (95% CI) | | | | |
| DenseNet-121 | 0.85 (0.85-0.86) | 0.86 (0.85-0.87) | 0.86 (0.86-0.87) | 0.86 (0.85-0.86) | 0.87 (0.86-0.87) |
| CXR-Linear | 0.80 (0.79-0.81) | 0.82 (0.81-0.83) | 0.80 (0.79-0.80) | 0.80 (0.79-0.80) | 0.82 (0.81-0.82) |
| CXR-MLP-3 | 0.81 (0.80-0.81) | 0.82 (0.81-0.83) | 0.80 (0.80-0.81) | 0.80 (0.80-0.81) | 0.82 (0.82-0.82) |
| CXR-MLP-5 | 0.81 (0.80-0.81) | 0.82 (0.81-0.82) | 0.79 (0.79-0.80) | 0.80 (0.79-0.80) | 0.82 (0.81-0.82) |
| | TPR (95% CI) | | | | |
| DenseNet-121 | 0.75 (0.73-0.76) | 0.73 (0.72-0.74) | 0.81 (0.80-0.82) | 0.75 (0.74-0.76) | 0.79 (0.78-0.80) |
| CXR-Linear | 0.61 (0.59-0.62) | 0.65 (0.64-0.66) | 0.69 (0.68-0.71) | 0.66 (0.65-0.67) | 0.66 (0.65-0.67) |
| CXR-MLP-3 | 0.62 (0.60-0.63) | 0.64 (0.63-0.65) | 0.71 (0.70-0.72) | 0.67 (0.66-0.68) | 0.66 (0.65-0.67) |
| CXR-MLP-5 | 0.61 (0.59-0.62) | 0.63 (0.62-0.64) | 0.68 (0.66-0.69) | 0.65 (0.64-0.66) | 0.64 (0.63-0.65) |
| | FPR (95% CI) | | | | |
| DenseNet-121 | 0.18 (0.18-0.19) | 0.17 (0.17-0.18) | 0.25 (0.24-0.25) | 0.19 (0.19-0.20) | 0.20 (0.20-0.21) |
| CXR-Linear | 0.17 (0.16-0.17) | 0.19 (0.18-0.19) | 0.25 (0.25-0.26) | 0.22 (0.22-0.22) | 0.18 (0.18-0.19) |
| CXR-MLP-3 | 0.18 (0.17-0.18) | 0.18 (0.17-0.18) | 0.25 (0.25-0.25) | 0.22 (0.22-0.22) | 0.18 (0.18-0.19) |
| CXR-MLP-5 | 0.18 (0.17-0.18) | 0.18 (0.18-0.19) | 0.24 (0.24-0.25) | 0.22 (0.21-0.22) | 0.19 (0.18-0.19) |
| | Youden's J statistic (95% CI) | | | | |
| DenseNet-121 | 0.57 (0.55-0.58) | 0.56 (0.54-0.57) | 0.56 (0.55-0.57) | 0.56 (0.55-0.57) | 0.58 (0.57-0.59) |
| CXR-Linear | 0.44 (0.43-0.46) | 0.46 (0.45-0.48) | 0.44 (0.43-0.45) | 0.44 (0.43-0.45) | 0.47 (0.46-0.48) |
| CXR-MLP-3 | 0.44 (0.42-0.45) | 0.46 (0.45-0.48) | 0.46 (0.45-0.47) | 0.45 (0.44-0.46) | 0.48 (0.47-0.49) |
| CXR-MLP-5 | 0.43 (0.42-0.45) | 0.45 (0.43-0.46) | 0.43 (0.42-0.44) | 0.43 (0.42-0.45) | 0.45 (0.44-0.46) |

Note.—Disease detection results reported separately for each race group and biological sex. TPR and FPR in subgroups are determined using a fixed decision threshold optimized over the whole patient population for a target FPR of 0.20. Performance is reported as mean and 95% confidence interval (CI). AUC = area under the receiver operating characteristic curve, TPR = true positive rate, FPR = false positive rate, CXR = chest radiography, MLP = multi-layer perceptrons

Table S4. Comparison of disease detection performance for the label 'pneumothorax'

| | White | Asian | Black | Female | Male |
|---|---|---|---|---|---|
| **Pneumothorax** | | | | | |
| Model | AUC (95% CI) | | | | |
| DenseNet-121 | 0.85 (0.84-0.86) | 0.87 (0.86-0.87) | 0.87 (0.86-0.88) | 0.87 (0.86-0.87) | 0.87 (0.86-0.87) |
| CXR-Linear | 0.81 (0.80-0.81) | 0.82 (0.81-0.83) | 0.83 (0.82-0.84) | 0.83 (0.82-0.83) | 0.82 (0.81-0.82) |
| CXR-MLP-3 | 0.80 (0.80-0.81) | 0.82 (0.81-0.83) | 0.84 (0.83-0.85) | 0.82 (0.81-0.83) | 0.83 (0.82-0.83) |
| CXR-MLP-5 | 0.81 (0.80-0.81) | 0.82 (0.81-0.83) | 0.82 (0.80-0.83) | 0.82 (0.81-0.82) | 0.82 (0.81-0.83) |
| | TPR (95% CI) | | | | |
| DenseNet-121 | 0.77 (0.75-0.78) | 0.78 (0.77-0.80) | 0.77 (0.75-0.79) | 0.77 (0.76-0.78) | 0.78 (0.77-0.79) |
| CXR-Linear | 0.67 (0.65-0.68) | 0.71 (0.69-0.72) | 0.69 (0.66-0.71) | 0.67 (0.66-0.69) | 0.70 (0.69-0.72) |
| CXR-MLP-3 | 0.69 (0.67-0.71) | 0.70 (0.68-0.71) | 0.67 (0.65-0.70) | 0.64 (0.63-0.67) | 0.72 (0.71-0.73) |
| CXR-MLP-5 | 0.68 (0.67-0.70) | 0.71 (0.69-0.72) | 0.67 (0.64-0.69) | 0.66 (0.64-0.68) | 0.71 (0.70-0.73) |
| | FPR (95% CI) | | | | |
| DenseNet-121 | 0.22 (0.22-0.23) | 0.21 (0.20-0.21) | 0.17 (0.17-0.17) | 0.18 (0.18-0.19) | 0.21 (0.21-0.22) |
| CXR-Linear | 0.22 (0.22-0.23) | 0.22 (0.21-0.22) | 0.16 (0.16-0.16) | 0.18 (0.18-0.19) | 0.21 (0.21-0.22) |
| CXR-MLP-3 | 0.22 (0.22-0.22) | 0.22 (0.22-0.22) | 0.16 (0.16-0.17) | 0.18 (0.17-0.18) | 0.22 (0.22-0.22) |
| CXR-MLP-5 | 0.22 (0.21-0.22) | 0.22 (0.22-0.23) | 0.16 (0.16-0.16) | 0.18 (0.18-0.18) | 0.22 (0.21-0.22) |
| | Youden's J statistic (95% CI) | | | | |
| DenseNet-121 | 0.55 (0.53-0.56) | 0.58 (0.56-0.59) | 0.60 (0.58-0.62) | 0.59 (0.57-0.60) | 0.56 (0.55-0.58) |
| CXR-Linear | 0.44 (0.43-0.46) | 0.49 (0.48-0.50) | 0.53 (0.50-0.55) | 0.49 (0.47-0.50) | 0.49 (0.48-0.50) |
| CXR-MLP-3 | 0.47 (0.45-0.49) | 0.48 (0.46-0.49) | 0.50 (0.48-0.54) | 0.47 (0.45-0.49) | 0.50 (0.49-0.51) |
| CXR-MLP-5 | 0.47 (0.45-0.48) | 0.48 (0.47-0.50) | 0.51 (0.48-0.53) | 0.48 (0.46-0.49) | 0.50 (0.48-0.51) |

Note.—Disease detection results reported separately for each race group and biological sex. TPR and FPR in subgroups are determined using a fixed decision threshold optimized over the whole patient population for a target FPR of 0.20. Performance is reported as mean and 95% confidence interval (CI). AUC = area under the receiver operating characteristic curve, TPR = true positive rate, FPR = false positive rate, CXR = chest radiography, MLP = multi-layer perceptrons